\documentclass[sigconf,nonacm]{acmart}
\usepackage{graphicx} 
\usepackage{xcolor}
\usepackage{multirow}
\usepackage{xspace}

\newcommand{\zk}[1]{\textcolor{black}{#1}}
\newcommand{\name}{SOLARIS\xspace}

\setcopyright{none}
\copyrightyear{}
\acmYear{}
\acmDOI{}
\acmISBN{}
\acmConference[]{}{}{}
\acmBooktitle{}
\acmPrice{}

\begin{document}

\title{SOLARIS: Speculative Offloading of Latent-bAsed Representation for Inference Scaling}

\author{Zikun Liu, Liang Luo, Qianru Li, Zhengyu Zhang, Wei Ling, Jingyi Shen, Zeliang Chen, Yaning Huang, Jingxian Huang, Abdallah Aboelela, Chonglin Sun, Feifan Gu, Fenggang Wu, Hang Qu, Huayu Li, Jill Pan, Kaidi Pei, Laming Chen, Longhao Jin, Qin Huang, Tongyi Tang, Varna Puvvada, Wenlin Chen, Xiaohan Wei, Xu Cao, Yantao Yao, Yuan Jin, Yunchen Pu, Yuxin Chen, Zijian Shen, Zhengkai Zhang, Jing Zhu, Dong Liang, Ellie Wen}
\affiliation{%
  \institution{Meta AI}
  \city{Menlo Park}
  \state{California}
  \country{USA}}

\begin{abstract}
Recent advances in recommendation scaling laws have led to foundation models of unprecedented complexity. While these models offer superior performance, their computational demands make real-time serving impractical, often forcing practitioners to rely on knowledge distillation—compromising serving quality for efficiency.

To address this challenge, we present \name (\textbf{S}peculative \textbf{O}ffloading of \textbf{L}atent-b\textbf{A}sed \textbf{R}epresentation for
\textbf{I}nference \textbf{S}caling), a novel framework inspired by speculative decoding. \name proactively precomputes user-item interaction embeddings by predicting which user-item pairs are likely to appear in future requests, and asynchronously generating their foundation model representations ahead of time. This approach decouples the costly foundation model inference from the latency-critical serving path, enabling real-time knowledge transfer from models previously considered too expensive for online use.

Deployed across Meta's advertising system serving billions of daily requests, \name achieves 0.67\% revenue-driving top-line metrics gain, demonstrating its effectiveness at scale.
\end{abstract}

\maketitle

\section{Introduction}
In recent years, foundation models (FMs) have revolutionized large-scale recommendation systems (RecSys), unlocking unprecedented modeling capacity, richer data utilization, and enhanced personalization. Modern FMs ~\cite{liang2025externallargefoundationmodel,luo2025metalatticemodelspace,zhang2024wukongscalinglawlargescale} excel at learning complex user-item interactions from massive, heterogeneous datasets, and consistently achived performance gains in offline evaluations.

Despite these advances, deploying FMs in production remains a major challenge. Their computational complexity makes real-time serving impractical, leading practitioners to rely on  knowledge transfer to smaller, vertical models (VMs). 

Unfortunately, the industry standard, knowledge distillation, faces two bottlenecks:

\begin{enumerate}
    \item \textbf{Limited Transfer Ratio and Generalizability}: soft-label based~\cite{hinton2015distilling,chen2017learning,tang2018ranking,kim2016sequence} distillation typically achieves only a 20--25\% transfer ratio and the resulting prediction labels are tightly coupled to specific tasks.
    \item \textbf{Limited Availability and Coverage}: Knowledge transfer occurs exclusively during training when labels are available~\cite{1_1,1_2,1_3}, while missing opportunities for knowledge sharing at inference time.
\end{enumerate}


To address these challenges, we propose \name (\textbf{S}peculative \textbf{O}ffloading of \textbf{L}atent-b\textbf{A}sed \textbf{R}epresentation for
\textbf{I}nference \textbf{S}caling)---a scalable, high-transfer-ratio, inference-time ready and production-friendly framework for large scale knowledge sharing in RecSys. \name introduces the following key innovations:

\begin{enumerate}
    \item \textbf{Direct Embedding-based Transfer}: \name moves beyond soft-label distillation by directly transferring user-item interaction embeddings from the foundation model to VMs. This approach yields richer, more generalizable representations and significantly better knowledge transfer ratio.
    \item \textbf{Speculative Embedding Precomputation}: Inspired by speculative decoding in LLMs~\cite{leviathan2023fastinferencetransformersspeculative}, \name anticipates which user-item interactions are likely to occur in the near future and precomputes their embeddings asynchronously. This enables inference-time knowledge distillation, allowing VMs to leverage foundation model representations during serving. 
    \item \textbf{Hierarchical Feature Enrichment}: To maximize coverage without incurring linear distillation costs, \name employs hierarchical feature enrichment. When a direct user-item match is unavailable, it aggregates user and item embeddings and similarity-based representations to continue providing knowledge transfer benefits.
\end{enumerate}


By decoupling the expensive knowledge transfer process from the latency-critical serving path, \name delivers richer user-item representations at inference time with no additional overheads.  \name seamlessly integrates with existing multi-stage ranking pipelines and scales to billions of daily requests, all while maintaining strict latency requirements in Meta's advertising infrastructure. 

Our extensive evaluation of \name in real world production environment demonstrates up to 0.2\% relative log loss (RLL) improvements on 10+ production models, lifting coverage by 30\%, and resulting in 0.67\% global ads revenue gain (on the order of \$100M).
\section{Background}
\begin{figure*}[t]
  \centering
  \includegraphics[width=\linewidth]{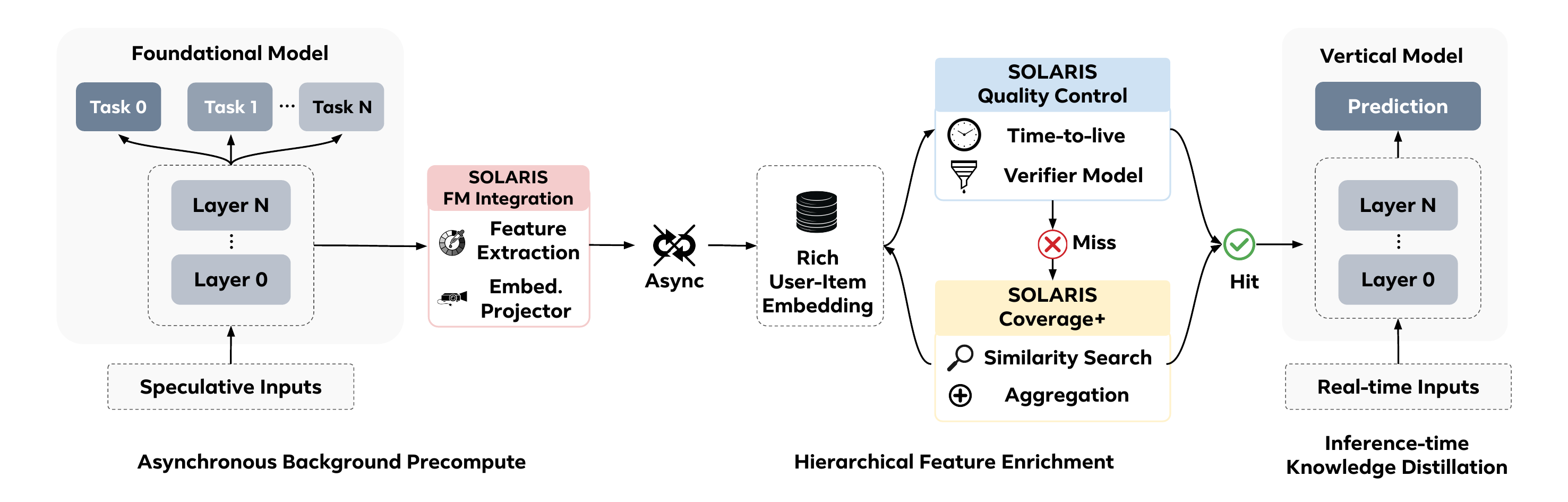}
  \caption{\name overview}
  \label{fig:polaris_overview}
  \vspace{-0.6cm}
\end{figure*}
\subsection{Multi-stage items Ranking}
Modern RecSys employs a multi-stage pipeline~\cite{b1,b2,b3} to efficiently select relevant items. The process consists of three stages: 1) Retrieval stage~\cite{b_r1,b_r2, b_r3}, where it filters items using broad, lightweight criteria such as user interests and ad targeting. 2) Early-stage ranking~\cite{b_e1,b_e2}, where it applies a lightweight deep learning model with user and ad features to narrow the selection to hundreds of items. 3) Final stage ranking~\cite{b_f1,b_f2,b_f3}, where it uses resource-intensive models that analyze thousands of signals, including real-time user activity, to select the top items for auction and delivery. In our system, \name serves the final stage ranking models.

\subsection{Knowledge Transfer}
Knowledge transfer is a fundamental technique in modern recommendation systems, enabling models to leverage information from data-rich domains or large teacher models to enhance performance in downstream or data-scarce scenarios. 

\noindent\textbf{Knowledge Distillation}: The teacher model provides soft labels—probability distributions over possible classes—to enable the student to learn not only the correct answers but also the relative similarities among classes ~\cite{hinton2015distilling,chen2017learning,tang2018ranking,kim2016sequence}. However, this method is inherently limited by the amount of information that can be conveyed through soft labels alone in ranking systems, typically lower than 25\%~\cite{tang2018ranking}. 

\noindent\textbf{Embedding Sharing}: Another widely used technique, where representations learned by upstream models are reused as features in downstream tasks~\cite{covington2016deep,cheng2016wide,shin2019pupil,es_1,es_2,es_3}. While effective, embedding sharing can introduce significant computational and storage overhead. 


\section{\name Design}
\subsection{System Overview}
Figure \ref{fig:polaris_overview} shows the overall \name architecture. The framework is built on the insight that foundation models (FMs) learn rich user-item interaction representations in their intermediate layers, which capture holistic knowledge from massive datasets, diverse features, and multi-objective optimization. 

\zk{In \name, we specifically extract embeddings from the last shared layer of the FM architecture. where concentrates the most relevant knowledge about user-item interactions. Such embedding computation happens asynchronously in the background and we designed the hierarchical feature enrichment module to obtain embeddings with boosted coverage and guaranteed freshness. These embeddings are used as input for downstream models to enhance performance.}  

\subsection{\name Knowledge Sharing}\label{sec: polaris_ks}
Traditional soft-label distillation~\cite{3_2_1,3_2_2,3_2_3} transfers knowledge only during training with limited efficacy. We design \name to address this fundamental limitation by enabling continuous knowledge transfer using a unified, FM-agnostic paradigm.


\subsubsection{Foundational Model Compatibility}~\label{sec:polaris_fm}
\zk{\name applies to all multi-task multi-label style FMs as shown in Figure ~\ref{fig:polaris_overview}. As an example, many SOTA FMs process heterogeneous input features through multiple parallel pathways~\cite{luo2025metalatticemodelspace, zeng2025interformereffectiveheterogeneousinteraction}: }(1) a dense architecture that transforms continuous features into low-dimensional embeddings via multi-layer perceptrons, (2) a sparse architecture that converts categorical features into pooled embeddings through embedding bag collections, and (3) a variable-length embedding architecture that captures sequential user behavior patterns from event-based features~\cite{sq_1,sq_2,sq_3,sq_4}. The outcomes of the three paths are then processed via a shared stack of interaction layers~\cite{zhang2024wukongscalinglawlargescale, liang2025externallargefoundationmodel, zhang2022dhen, wang2021dcn}, producing the shared network output. 



\subsubsection{\name Feature Extraction}  The user-item embeddings are derived from the output of our shared feature interaction network, which processes heterogeneous input features—including user demographics, behavioral sequences, and item attributes. This embedding is then compressed via an autoencoder~\cite{ae1,ae2} to produce a compact, fixed-dimensional vector suitable for downstream consumption. This embedding serves as additional input features to vertical models. This knowledge sharing mechanism enables vertical models to leverage the rich user-item interaction patterns learned from the FM trained across different domains.

\subsubsection{Knowledge Sharing During Inference Time} The rich embeddings produced by FMs are stored alongside other input features, and are consumed just like other features by the VMs, inculding during serving. And online training is conducted on FM to adapt to shifts in distribution.

\subsection{Asynchronous Precompute}\label{sec:async_polaris}
However, it is impractical to generate FM user-item interaction for each user-item pair from the user and items pool, due to prohibitive infrastructure compute and storage cost. Therefore, we propose to generate the embeddings selectively via asynchronous precompute.
At a high level, \name maintains a background service to compute user-item embeddings using a FM. It takes user-level requests, evaluates item candidates in those requests, and cache the generated user-item embeddings to be consumed by future requests. The async precompute process relaxed the latency budget by bringing embedding generation to the background. As a trade-off, this approach compromises feature freshness as well as comprehensive coverage at user-item level. 

\zk{\textbf{Background Teacher Computation:} When a user request comes in, multi-stage ranking phases narrow down the items candidates to hundreds. The FMs maintain background jobs that continuously generates the user-item embedding from top 20\% most relevant items to the user (decided by verifier model mechanism) from these candidates. The computed embeddings are stored in a distributed cache, keyed by <user, item> pairs.}

\textbf{Verifier Model Mechanism:} To further improve efficiency and cache quality, we employ a lightweight verifier model—specifically, the prediction score from the current vertical model—to decide whether a user-item pair should be retained or discarded from the cache. Only user-item pairs with sufficiently high predicted relevance are selected for embedding precomputation and caching. This mechanism acts as a real-time filter, ensuring that computational resources are focused on the most promising candidates, and aligns with the speculative decoding paradigm: we “speculate” which user-item pairs will be needed and discard the rest.

\textbf{TTL-based Cache Invalidation:} The success of asynchronous precomputation is contingent upon the stability of a user's preferences and interests within a short time frame. In early-stage ranking, since user preferences typically remain consistent within a few hours (it has been observed that over 60\% of item candidates in the current request were ranked within the past six hours), teacher embeddings retain their relevance, enabling effective vertical model enhancement during inference without requiring real-time teacher computation. During both training and inference, vertical models query the cache for each user-item pair to retrieve precomputed embeddings. The system enforces a cache reading TTL (typically a few hours) to ensure embedding freshness while balancing computational efficiency. 

\textbf{Vertical Model Serving:} As described in Section~\ref{sec: polaris_ks}, vertical models takes the user-item embeddings as additional input features to improve predictions. When a <user, item> query is missing or expired from the cache, the user-item pair is queued for the \name model's next refresh cycle while zero tensors serve as placeholders in the vertical models input.

\subsection{Hierarchical Feature Enrichment}\label{sec:coverage}
                While asynchronous precomputation enables efficient knowledge transfer, achieving complete coverage for all user-item pairs remains challenging due to the vast scale of the user and item space, and the infrastructure concerns. As a result, the vanilla user-item embedding coverage is approximately 50\% in our system. This coverage gap is particularly pronounced for long-tail items, new users, or emerging user-item interactions that haven't been prioritized in recent precomputation cycles. To address the coverage limitations, \name implements 2 complementary strategies that improves embedding availability while maintaining system efficiency.

\textbf{Aggregated user-item Embedding:} \name introduces an aggregated user-only embedding, computed by averaging all available user-item embeddings for a given user across different items within the past 24 hours, explicitly excluding the embedding for the current user-item pair. This approach leverages the assumption that user preferences exhibit short-term consistency, allowing recent interaction embeddings to inform current predictions. The aggregated user-item embedding serves as an additional input feature, supplementing the standard user-item embedding, and increases effective feature coverage to approximately 85--90\%, ensuring that most users have some form of foundation model representation available during inference.

\textbf{Similarity-Based Embedding: } \name maintains a precomputed user clustering table that captures user similarity relationships. The clustering process operates in two phases: first, a dedicated FM generates comprehensive user embeddings that encode rich behavioral representations; second, a k-nearest neighbors (KNN) algorithm identifies similar users for each target user based on cosine similarity in the embedding space. For each user-item pair lacking a precomputed embedding, \name queries the clustering table to identify the top similar users to the target user. Among these similar users, \name searches for available user-item embeddings of form <neighbor, items>. \name then employs one of two strategies for leveraging neighbor embeddings: (1) aggregating multiple neighbor embeddings using distance-weighted averaging, where closer users receive proportionally higher weights, or (2) selecting the embedding from the single most similar neighbor based on proximity in the user feature space. This similarity-based approach extends the foundation model's effective coverage by 30\%. The method leverages collaborative filtering principles, assuming that users with similar behavioral patterns will exhibit comparable responses to identical advertisements, thereby providing meaningful signal for previously unseen interactions.

\section{Experiments}
We now comprehensively evaluate \name with real-world serving scenarios, demonstrating its effectiveness in lifting serving quality without latency overhead. We then carfully ablate the contribution from individual \name components.

\subsection{Deployment Baseline and Methodology}
While our framework is applicable to general user-item scenarios, we demonstrate its effectiveness on Meta ads in our experiments. We have deployed \name across a wide range of advertising model types serving global users across platform. Our baseline is prior production system without \name, where each portfolio was served by a separate model trained on isolated datasets. During testing, both \name and the baseline predict CTR(Click-Through Rate)/CVR(Conversion Rate) metrics for live traffic in the form of (user, ad) pairs, with the highest-ranked pairs presented to users based on predictions and other factors.

\begin{table}[]
\centering
\resizebox{\linewidth}{!}{%
\begin{tabular}{c|c|c}
\hline
Task Categories      & Sub-tasks             & Relative BCE Loss $\downarrow$ (\%) \\ \hline
\multirow{4}{*}{CTR} & Facebook Feed         & 0.09                      \\
                     & Facebook Reels        & 0.08                      \\
                     & Instagram             & 0.05                      \\
                     & Instagram Link Click  & 0.05                      \\ \hline
\multirow{2}{*}{CVR} & Facebook Feed + Reels & 0.1                       \\
                     & Offsite Conversion    & 0.05                      \\ \hline
\end{tabular}%
}
\caption{\name Embedding Improves Vertical Model Performance on Varieties of Tasks}
\label{tab:vertical}
\vspace{-0.6cm}

\end{table}

\subsection{Results on Vertical Models}
Table \ref{tab:vertical} summarizes the impact of integrating \name user-ad embeddings on a diverse set of CTR and CVR tasks across Meta’s major ad products. Here, we report the relative reduction in binary cross-entropy (BCE) loss (\%) for each sub-task, reflecting improvements in model prediction quality. Sub-tasks are defined as distinct prediction objectives within each task category, corresponding to specific surfaces or user actions (e.g., Facebook Feed CTR, Instagram Link Click CTR, Offsite Conversion).

\noindent\textbf{CTR Gains: }\name user-ad embeddings consistently improved CTR prediction across all evaluated surfaces. Specifically, we observed a 0.09\% reduction in BCE loss for Facebook Feed, 0.08\% for Facebook Reels, and 0.05\% for both Instagram Feed and Instagram Link Click sub-tasks. These improvements indicate that embedding sharing enables the model to better capture user-ad interaction patterns, leading to more accurate click probability estimates. Notably, the gains are present not only in primary CTR tasks (e.g., Feed, Reels) but also in auxiliary tasks such as Instagram Link Click, demonstrating the generality of the approach.

\noindent\textbf{CVR Gains: }For conversion rate tasks, \name led to clear reductions in prediction loss and higher conversion rates across both on-platform and offsite events. The model achieved a 0.10\% reduction in BCE loss for the combined Facebook Feed + Reels conversion task, and a 0.05\% reduction for offsite conversion (e.g., purchases or sign-ups occurring outside Meta platforms). These results highlight the effectiveness of embedding-based knowledge transfer in improving conversion prediction, especially in scenarios where conversion signals are sparse or user intent is more difficult to model, showing the value of \name for downstream business objectives.

\noindent\textbf{Transfer Ratio: }\name achieves a transfer ratio of 42\% on Instagram tasks and 44\% on Facebook products, representing a 2X increase over previous distillation-based approaches. This shows the effectiveness of \name's unique knowledge sharing.

\begin{table}[t]
\centering
\resizebox{1\linewidth}{!}{ 
\begin{tabular}{l|l|l|l|l}
\hline
\name Feature Coverage                   & 20\%   & 50\%  & 60\%   & 100\% \\ \hline
BCE $\downarrow$  on Instagram CTR & 0.05\% & 0.1\% & 0.13\% & 0.3\% \\ \hline
\end{tabular}
}
\caption{\name Feature Coverage Study}
\vspace{-1cm}

\label{tab:feature}
\end{table}

\subsection{\name Embedding Coverage Results}
\textbf{Feature Coverage VS Performance: } As described in Sec~\ref{sec:async_polaris},  \name adopts asynchronous pre-computation to generate user-ad embeddings during serving time. We deployed \name in production environment and user-ad embedding shows coverage of 40\%. The coverage is calculated on the training data where the labels (click, conversion, etc) are generated.  We conducted the offline experiment on user-ad embedding coverage's effect on downstream models' performance boost. The offline user-ad embeddings are generated by the same \name model but during offline training and evaluation so that the coverage achieves nearly 100\%. We down-sample the user-ad embeddings to achieve different coverages and add to the downstream model.  As shown in Table~\ref{tab:feature}, when the feature coverage increases by 30\%, the log loss is improved by 0.05\% which is considered significant on our production models. This shows great potential for unblocking more ads revenue when the \name feature coverage increases.

\noindent\textbf{Aggregated User-ad Performance: } As described in Sec~\ref{sec:coverage}, we introduces an aggregated version of user-ad embedding as user-only embedding, leading to the coverage increases to 90\%. During production deployment, we deploy user-only embedding as a separate feature to achieve 0.03\% additional BCE loss improvement on Instagram CTR tasks. This user-only feature has been deployed to 10 downstream models in real-world production pipelines and increases ads revenue by 0.07\%.

\noindent\textbf{Similarity-based Embedding Performance: } We deployed an KNN operator to obtain 100 similar neighbors for each user and impute the missing <user, ad> embedding from the most similar neighbor's user-ad embedding, leading to the coverage of 70\% compared to 40\% before. Offline experiments on Instagram CTR task shows the BCE loss improving of 0.02\%. Compared to our feature coverage study, this achieves 40\% of the headroom (0.02\%/0.05\%). This is as expected since the imputed user-ad embeddings from neighbors have lower quality.
\section{Discussion and Limitation}
\noindent\textbf{Further Enhancing Coverage: } The current clustering method for improving coverage yields only limited performance gains on vertical models. Our ongoing efforts will expand beyond the U2U (user-to-user) approach described in this paper to also explore A2A (ad-to-ad) and hybrid strategies. 

\noindent\textbf{Early-stage Roll Out: } \name is only operational and beneficial for vertical models in the final ranking stage. This limitation arises because earlier ranking stages involve 100X more ads, making coverage and resource constraints significant challenges. Given the vast model space in these early stages, we are actively working to increase embedding coverage, aiming to unlock greater revenue potential.

\noindent\textbf{Limitations: } While we aim to share broadly applicable deployment insights, some findings are inherently company-specific. For instance, business requirements and partner relationships at Meta shape the problem spaces addressed by \name, which may limit the generalizability of certain results.

\balance
\bibliographystyle{ACM-Reference-Format}



\end{document}